\documentclass[conference]{IEEEtran}
\IEEEoverridecommandlockouts
\pdfoutput=1

\usepackage[super]{nth}
\usepackage{graphicx}
\usepackage{textcomp}
\usepackage[table, x11names]{xcolor}
\usepackage{multirow}
\usepackage{placeins}
\usepackage{bm}
\usepackage{booktabs}
\usepackage{caption}
\usepackage{hyperref}
\usepackage{amsmath,amssymb,amsfonts}
\usepackage{mathtools}

\usepackage{algorithmic}
\usepackage{xfrac}

\usepackage{graphicx}
\usepackage{subfig}
\usepackage{float}

\usepackage{cite}
\graphicspath{ {plots/} }

\def\BibTeX{{\rm B\kern-.05em{\sc i\kern-.025em b}\kern-.08em
    T\kern-.1667em\lower.7ex\hbox{E}\kern-.125emX}}
    
\begin{document}

\title{Modelling Early User-Game Interactions for Joint Estimation of Survival Time and Churn Probability\\
\thanks{This work was supported by the EPSRC Centre for Doctoral Training in Intelligent Games \& Games Intelligence (IGGI) [EP/L015846/1] and the Digital Creativity Labs (digitalcreativity.ac.uk), jointly funded by EPSRC/AHRC/Innovate UK under grant no. EP/M023265/1.}
}

\author{
\IEEEauthorblockN{V. Bonometti\dag\ddag, C. Ringer\dag*, M. Hall\ddag, A.R. Wade\S, A. Drachen\dag}
\IEEEauthorblockA{\dag Department of Computer Science, University of York, York, UK YO10 5DD\\
*Department of Computing, Goldsmiths, University of London, London, UK SE14 6NW\\
\ddag Square Enix Limited, London, UK SE1 8NW\\
\S Department of Psychology, University of York, York, UK YO10 5DD \\
Email: \{vb690, cr1116, anders.drachen, alex.wade\}@york.ac.uk; analyticsl@eu.square-enix.com}
}

\maketitle

\begin{abstract}
Data-driven approaches which aim to identify and predict player engagement are becoming increasingly popular in games industry contexts. This is due to the growing practice of tracking and storing large volumes of in-game telemetries coupled with a desire to tailor the gaming experience to the end-user's needs. These approaches are particularly useful not just for companies adopting Game-as-a-Service (GaaS) models (e.g. for re-engagement strategies) but also for those working under persistent content-delivery regimes (e.g. for better audience targeting). A major challenge for the latter is to build engagement models of the user which are data-efficient, holistic and can generalize across multiple game titles and  genres with minimal adjustments. 

This work leverages a theoretical framework rooted in engagement and behavioural science research for building a model able to estimate engagement-related behaviours employing only a minimal set of game-agnostic metrics. Through a series of experiments we show how, by modelling early user-game interactions, this approach can make joint estimates of long-term survival time and churn probability across several single-player games in a range of genres. The model proposed is very suitable for industry applications since it relies on a minimal set of metrics and observations, scales well with the number of users and is explicitly designed to work across a diverse range of titles. Code is available at \url{https://github.com/vb690/churn_survival_joint_estimation}.
\end{abstract}


%%%%%%%%%%%%%%%%%%%%%%%%%%%%%%%%%%%%%%%%%%%%%%%%%%%%%%%%%%%%%%%%%%%%%%%%%%%%%%%%%%%%%%%%%%%%%%

\section{Introduction}
The video game industry has gained the ability to draw insights on the playing activity from extremely large cohorts of users. This is made possible by the increasing practice of recording a wide range of in-game telemetries coupled with the possibility to store and process massive amounts of data. One of the most important use-case for this type of data is to develop solutions for assessing and predicting engagement-related behaviours \cite{milovsevic2017early,drachen2016rapid,runge2014churn,hadiji2014predicting}. Indeed, given that the ultimate goal of a game is to deliver a specific entertaining experience to the end user, understanding if, how, and when players are engaged with a game has a pivotal importance in many different areas of applications such as: play-testing \cite{mirza2013does}, stakeholder reporting \cite{drachen2016rapid} and automated in-game behaviour estimation \cite{milovsevic2017early}. This last one has become perhaps one of the most important goal in contemporary Game Analytics with a major focus on tasks such as churn and survival estimation. Despite the literature on the topic being relatively narrow, a clear pattern of collaboration between academia and industry seems to emerge where one tries to propose solution useful for the other \cite{milovsevic2017early,runge2014churn,hadiji2014predicting,drachen2016rapid,viljanen2018playtime}. In this view, the present work tries to close some of the gaps in the aforementioned literature proposing a novel `industry-friendly' approach for estimating engagement-related behaviours. We designed a model for joint estimation of survival time and churn probability working across a range of games contexts. Our model requires minimal pre-processing, it employs a restricted set of game-agnostic features, it is suitable for large scale applications and it is able to incorporate uncertainty in its estimations. Moreover, through a series of 3 experiments, we showed how employing a `hybrid approach' \cite{yannakakis2013player} and integrating insights from the engagement and behavioural science literature into the model design allowed us to achieve consistently better results compared to baseline and competing approaches.

%%%%%%%%%%%%%%%%%%%%%%%%%%%%%%%%%%%%%%%%%%%%%%%%%%%%%%%%%%%%%%%%%%%%%%%%%%%%%%%%%%%%%%%%%%%%%%%%%%%%%%%%%%%%%%%%%%%%%%%%%%%%%%%%%%%%%%%%%%%%%%

\section{Engagement} \label{Related work: Engagement}
Due to space constrains, this section cannot present an extensive critical review of engagement, indeed our aim is not to provide a theoretical contribution to the construct but rather to operationalize it from a behavioural point of view. Highlighting connections between engagement and behavioural science, we tried to develop a guiding framework for the modelling attempts carried out in our experiments.

\subsection{A video-game perspective on engagement}
%\subsection{A video-game perspective}
Defining engagement in digital games is a non-trivial task, with past attempts relying on constructs such as: Flow \cite{sherry2004flow}, Self Determination Theory \cite{ryan2006motivational}, Immersion \cite{jennett2008measuring} and Uses and Gratification Theory \cite{lucas2004sex}. However, due to variation in theoretical formulations and the scarcity of empirical validation, the underlying framework appear to be too heterogeneous and not strictly formalized \cite{boyle2012engagement}. Therefore, in the present work we decided to adopt a conceptualization, proposed by O'Brien and Toms \cite{o2008user}, which we believed formalized engagement in a more precise and most importantly operationalizable way. The authors describe engagement as a multiphasic process arising from the continuous interaction between the user and the game. Three stages of the aforementioned process are particularly relevant to this work: point of engagement, period of engagement and disengagement. The point of engagement is the moment in which a user directs their attention towards a game due to its ability to fulfill specific drives the user has. The period of engagement is defined by the sustained interaction of a user with a game and is maintained by the ability of the game to provide satisfying feedback. Finally, disengagement defines the moment in which a user makes an active decision to stop interacting with a game due to either external factors or the inability of the game to keep providing rewarding experiences. 

Summarizing, engagement seems to be a dynamic process in which the interaction between user and game has a central role. Moreover, this process seems to be initiated, sustained and terminated in accordance with the capability of the user-game interaction to provide rewarding experiences, as highlighted also by \cite{wang2011game, phillips2013videogame, avserivskis2017computational}.

\subsection{A behavioural perspective on engagement}
Considering the previous section, we could argue one way to model the state of the user during the engagement process could be to assess whether they are receiving a sufficiently rewarding experience. This becomes a complex task when performed on a large scale and without the direct involvement of the user because it requires assessing latent, non-observable, states (i.e. the experience of rewarding consequences) influencing measurable outcomes (i.e. behaviour). We can adopt some simple yet powerful concepts from behavioural science (i.e. operant conditioning) to assess whether an activity has reinforcing consequences for an individual by observing the individual's behaviour during the interaction between the two. A comprehensive discussion on `operant conditioning' would be impossible here given space constraints but, in principle, we say that a specific behaviour tends to increase in amount, frequency and duration when precise reinforcing (i.e. rewarding) consequences are associated to it \cite{skinner1953science, berridge2009dis}. 

Re-framing this in a video game context we could hypothesize that, in absence of external interference, a user will produce in-game behaviour for as long as the game is able to provide rewarding experiences. On the contrary if the game environment fails to provide these experiences, the user would reduce the amount and frequency of in-game behaviour until eventually complete disengagement occurs.

\subsection{Survival Time and Churn Probability as Engagement Approximation}
Engagement, while a complex construct, must be circumscribed to simple and quantifiable behaviours when employed for data-driven applications. For this reason we propose survival time and churn as behavioural approximations of future sustained engagement and disengagement. Generally speaking, survival time can be defined as the amount of playing activity occurring between the end of an observation period and the last activity recorded for a specific user \cite{ perianez2016churn, demediuk2018player, bertens2017games, kim2017churn, viljanen2018playtime}. Churn can be defined as the decision of a user to stop interacting with a specific service due to internal or external reasons, usually formalized as a user entering a prolonged period of inactivity \cite{hadiji2014predicting,runge2014churn, drachen2016rapid,milovsevic2017early, kim2017churn}. While GaaS can only rely on an inactivity period for determining churn, titles with a defined life cycle (e.g. AAA single player games) can utilize a defined end-game period as a hard cut-off for distinguishing between churners and non-churners: users finishing a game are not churners even if they stop playing afterwards. 

In summary, a model estimating survival time and churn from the early stages of the engagement process should include a number of characteristics: 1) Integrate the user-game interaction from the point of engagement through to the end of the on-boarding phase \cite{mirza2013does}, an initial period of sustained engagement critical for assuring long-term engagement; 2) Integrate in-game metrics indicative of behavioural activity to try to infer to what extent the game is providing rewarding experiences; 3) Explicitly model temporality since engagement appears to be a dynamic process which develops over time.

%%%%%%%%%%%%%%%%%%%%%%%%%%%%%%%%%%%%%%%%%%%%%%%%%%%%%%%%%%%%%%%%%%%%%%%%%%%%%%%%%%%%%%%%%%%%%%
\section {State of the art and Current Contribution}
\subsection{Churn and Survival Estimation}
Given space restrictions, it is not possible to exhaustively describe all the works related to survival and churn estimation, we will therefore focus on some key examples closely linked to our work. When it comes to estimating churn, in particular for industry applications, it is relevant to develop and test models considering different titles and genre. In this view \cite{runge2014churn, hadiji2014predicting, xie2015predicting, kim2017churn} are notable examples where a range of different modelling techniques (i.e. Linear Models, Decision Tress, Naive Bayes, Support Vector Machines and Deep Neural Networks (DNN)) were tested on a churn estimation task across multiple game titles. However, they often employed game specific features (sometimes carefully engineered) and build and test separate models for each game. A notable exception to that is the work done by \cite{liu2018semi}, where churn was formalized as edge prediction in a dynamic graph and modelled through a DNN. This produced a single model able to generalize across multiple game titles however with the limitation of them being all mobile titles. Another important characteristics for a churn estimation model, is to be able to produce predictions even when minimal observations are provided. In this view the works done by \cite{drachen2016rapid, milovsevic2017early} highlights the effort made in the literature for designing methodologies able to rapidly provide estimations of churn probability via minimal amount of metrics, this was done employing traditional machine learning algorithms (same as above) but exclusively considering metrics recorded during the initial stages of the user-game interaction. One of the major drawbacks in these works was that the initial period of observation was arbitrarily chosen and fixed for all the considered users making it difficult to take inter individual differences into account. In regard to the literature on survival analysis, we found that most works employed Cox Regression \cite{cox1972regression}, or some variation of it, for estimating the probability to survive (i.e. not have churned) after a specific period of time \cite{perianez2016churn, bertens2017games, demediuk2018player}. Despite being a similar formulation, this is not equivalent to estimating the survival time (i.e. the amount of future playing time), which becomes much more interesting when trying to assess not only measures of disengagement but also measures of future sustained engagement. A notable exception to this is the churn and survival analysis competition presented by \cite{liu2018semi}, where the goal was to estimate both churn probability and survival time, this also highlights the growing interest for richer assessments of user engagement and for models able to perform both tasks. On top of what is illustrated so far, we also individuated a series of limitations regarding the employed data-sources. The number of the considered users rarely goes beyond $10^4$ \cite{liu2018semi} and when it comes to churn estimation the class distribution is usually greatly imbalanced, both of these factors can pose limitations on the interpretation and generalization of results.

\subsection{Aims and contributions of the present work}
The contributions of the present work are two-fold: 1) The three experiments carried out provide insights on the validity of hybrid approaches in engagement modelling while trying to create a bridge between theoretical formulations and data-driven application. 2) The new modelling approach presented tries to fill some of the aforementioned gaps in the literature maintaining at the same time characteristics that makes it appealing for industry applications. While in previous works estimating survival time and churn probability was handled by separate models, our approach can perform both tasks in conjunction, providing a more holistic assessment of user engagement. The model was validated across four separate game genres, deviating from previous works which focused on a single game or a single game genre, potentially limiting the generalizability of the results. Furthermore, in contrast with previous works often utilizing large sets of human-engineered or game-specific features, our proposed model employs only a minimal collection of almost unprocessed and completely game-agnostic metrics, factor that can help reduce overhead in model deployment. In addition, we included in our model the capability to incorporate uncertainty in the predictions, allowing for more cautious interpretation of its estimates when employed in production pipelines.

%%%%%%%%%%%%%%%%%%%%%%%%%%%%%%%%%%%%%%%%%%%%%%%%%%%%%%%%%%%%%%%%%%%%%%%%%%%%%%%%%%%%%%%%%%%%%%

\section{Methodology}

\begin{table*}[h] \centering
\caption{\textbf{Data-set Description}. For each game we retrieved 80,000 Churners and 80,000 Non-Churners randomly sampled from all the available users.}
\label{gamesdescription}
\begin{tabular}{@{}lrrrrrrl@{}}
\toprule

\multirow{2}{*}{\textbf{Game}} & \multicolumn{2}{l}{\textbf{Survival Time (Mins})} & \multirow{2}{*}{\textbf{Churners}} & \multirow{2}{*}{\textbf{Non Churners}} & \multicolumn{2}{l}{\textbf{Observation Period}} & \multirow{2}{*}{\textbf{Descriptive Tags}} \\ \cmidrule(lr){2-3} \cmidrule(lr){6-7}
                      & \textbf{Min}                  & \textbf{Max}                  &                           &                               & \textbf{Min}                & \textbf{Max}               &                                \\ \midrule
hmg                        & 11 & 260    & 80,000 & 80,000  & 1  & 7  & Mobile, Single Player, Strategy                       \\
hms                        & 2 & 454     & 80,000 & 80,000  & 1  & 15 & Mobile, Single Player, Shooting Gallery                \\
jc3                        & 32 & 12,695 & 80,000 & 80,000  & 1  & 20 & Console \& PC, Single Player, Open World, Action              \\
jc4                        & 7 & 1,135   & 80,000 & 80,000  & 1  & 9  & Console \& PC, Single Player, Open World, Action              \\
lis                        & 5 & 704     & 80,000 & 80,000  & 1  & 6  & Console \& PC, Single Player, Story Driven, Graphic Adventure \\
libf                       & 14 & 1,214  & 80,000 & 80,000  & 1  & 10 & Console \& PC, Single Player, Story Driven, Graphic Adventure \\ \bottomrule
\end{tabular}
\end{table*}

\subsection{Data}
To conduct our experiments, we gathered data from six different games published by our partner company, \textit{Square Enix Limited}. Focusing on maintaining heterogeneity in genre and platform, we considered the following titles: \emph{Hitman Go} (hmg), \emph{Hitman Sniper} (hms), \emph{Just Cause 3} (jc3), \emph{Just Cause 4} (jc4), \emph{Life is Strange} (lis), and \emph{Life is Strange: Before the Storm} (lisbf). A general description of each of these titles can be found in Table \ref{gamesdescription}. Data were gathered from any user playing between the game's release and February 2019, allowing us to adopt more robust sampling strategies which utilizes the breadth of virtually the entire user-base. To rule out possible `faulty' but not `naturally abnormal' data, we restricted the data cleaning process to a single filter applied at query time to ignore users having at least one of the considered metric over the game population's \nth{99} percentile. This allowed us to make little assumptions on the distribution of the data as well as providing a convenient stress test for eventual future applications.

\subsubsection{Defining the Observation Period}
Because we were interested in estimating survival time and churn probability based only on early user-game interactions it was important to define a cut-off at which point interactions were no longer be considered `early'. We call the period from the user's first interaction till this cut-off the observation period (OP). Choosing the length for the OP was not trivial as there is little indication in the literature about optimal cut-off values. Hence, we decided to visually inspect the data a-priori and extend rules proposed in \cite{drachen2016rapid, milovsevic2017early} to take into account natural inter-individual differences. Therefore, we defined the cut-off as:

\begin{equation}
\label{CutoffOP}
    \text{cutoff} = 
    \Biggl\lceil
        \dfrac
            {min(S_t, S_c)}
            {3}
    \Biggr\rceil
\end{equation}

Where $S_t$ is the total number of game play sessions and $S_c$ is the number of game play sessions before the user completed the game for the first time. In this way we take the first \sfrac{1}{3} of all played sessions for players who churned and the first \sfrac{1}{3} of played sessions before a non-churning player completed the game for the first time. We apply this cut off to the ordered list of all recorded play sessions for a specific user. We decided to use game sessions as the temporal dimension, rather than total minutes played, since we believed it better adjusted for each user's `pace' (i.e. not all the users have the possibility to play at the same frequency). Since the length of the OP has a naturally different distribution between the churning and non-churning population, we stratified our sampling technique to maintain a similar ratio of OP lengths among churners and non churners. This becomes particularly relevant for Experiment 2 and 3 where the length of the OP could leak information in the churn probability estimation task. Summarizing, if a user for example had 9 total sessions recorded, we considered the first 3 for making estimations on what happened after the 9$^{th}$. It goes without saying that at production time the OP is defined only for generating the training samples, the model can be deployed at various stages of previously unseen time series which we simulate in our experiments with the test set. 

\subsubsection{Defining the Behavioural Metrics and Targets}
We considered a set of 5 metrics, easily generalizable across games and indicative of behavioural activity, and retrieved them temporally  (i.e. over each game session during the OP), see Table \ref{metricsdescription} for a description. Additionally, we acquired a single context feature specifying the game context from where the metrics were originated. For determining the targets for our survival and churn estimation tasks, we leveraged existing literature on churn prediction \cite{drachen2016rapid, milovsevic2017early, lee2018game, perianez2016churn, runge2014churn, kim2017churn, hadiji2014predicting, xie2015predicting} and survival analysis \cite{viljanen2018playtime, demediuk2018player, lee2018game, bertens2017games}, extending existing rules to accommodate the need to define churn and survival time in single player games with a defined life cycle (i.e. non-GaaS games). We took advantage of having access to the complete session history for all users to create a churn definition which was robust to the variance in play patterns across games, as it takes into account all the recorded inter-session distances. Therefore, the criteria we adopted for defining a user as churner were both: 

\begin{enumerate}
    \item Not completing the game
    \item Being inactive for a period equal or greater to:
        \begin{equation}
            \label{inactivityrule}
            \text{inactivity} = 
            mean(\mathbf{x}) + 2.5 \cdot std(\mathbf{x})
        \end{equation}
\end{enumerate}

For better adjusting for inter-individual differences, we could have applied formula \ref{inactivityrule} to each user individually but this could have created accuracy issue for individuals with very few recorded sessions. Therefore, we opted for a conservative but more robust approach applying inactivity ($\mathbf{x}$) $\forall \mathbf{x} \in X$ where $X$ is the collection of all the considered games and $\mathbf{x}$ is the vector of inter-sessions distances in minutes for a specific game. The use of formula \ref{inactivityrule} allowed us to estimate an inactivity period which was not arbitrarily chosen but statistically defined as ‘extraordinary long’ in accordance with characteristics of play patterns in a particular game. For defining the survival time, we simply computed the total amount of Play Time in minutes for a user minus the amount of Play Time during the OP.

\begin{table}[h] \centering
\caption{\textbf{Considered Metrics over Sessions}}
\label{metricsdescription}
\resizebox{0.5\textwidth}{!}{
\begin{tabular}{@{}ll@{}}
\toprule
\textbf{Metric}            & \textbf{Description}                   \\ \midrule
{Session Time}         & Overall session duration (minutes)              \\ 
{Play Time}            & Session Time spent actively playing (minutes)    \\ 
{Delta Session}        & Temporal distance  between sessions (minutes)   \\ 
{Activity Index}       & Count of user initiated game-play-related actions. E.g.\\ 
                       & `Talk to NPC' or `Acquire Upgrade' were considered valid\\ 
                       & actions while `Click Menu' or `NPC Attacks You' were not.\\
{Activity Diversity}   & Count of unique voluntarily initiated actions \\ 
{Context}              & Name of the game taken into consideration \\ \bottomrule
\end{tabular}
}
\end{table}

\subsubsection{Data Preparation}
We adopted specific data preparation procedures for each experiment. For the first analysis we collapsed the data over the temporal dimension retrieving mean and standard deviation of each considered features, to this concatenating a one-hot encoded transformation of the context metric. For the second and third experiments we kept the data in the original temporal form. In Experiment 3 only we treated the game context slightly differently, numerically encoding it and separating it from the other feature matrix. Since in Experiment 2 and 3 the length of the OP differed between users, we zero padded each sequence of considered sessions to the length of the longest sequence in the data-set. For each experiment we created a tuning and validation subsets (i.e. 20 and 80 \% of the original data-set) via stratified shuffle split \cite{scikit-learn}, employing the first for hyper-parameters searching and the second for model evaluation.

\subsection{Experiments}
For all experiments we applied the same procedure: first, determined the best hyper-parameters via grid search 10-fold stratified cross validation \cite{scikit-learn} on the tuning set then evaluated performance via 10-fold stratified cross validation on the validation set. In all experiments, we re-scaled the considered metric separately for each game in outliers-robust way, as in:

\begin{equation}
\label{robustscaler}
    \text{RobustRescale}=
        \dfrac
            {\mathbf{x} - Q_2(\mathbf{x})}
            {Q_3(\mathbf{x}) - Q_1(\mathbf{x})}
\end{equation}

where $\mathbf{x}$ is the feature vector to be re-scaled and $Q_n$ is the $n^{th}$ quartile for this game. The performance metric that we chose for our survival task was the Symmetric Mean Absolute Percentage Error (SMAPE), defined as:

\begin{equation}
\label{smape}
    \text{SMAPE}=
        \dfrac
            {\sum\limits_{i=1}^{N} \mid \hat{y_i} - y_i \mid}
            {\sum\limits_{i=1}^{N}  (y_i + \hat{y_i})}
\end{equation}

where $N$ is the collection of all the users in the considered set and $\hat{y_i}$ and $y_i$ are respectively estimated survival time and ground truth value for user \textit{i}. SMAPE was implemented because its scale invariance allowed better comparisons of results across game contexts. For the churn estimation task the chosen metric was the F1 score (F1), defined as:

\begin{equation}
\label{f1}
    \text{F1}=
        2 \cdot 
        \dfrac
            {(precision \cdot recall)}
            {(precision + recall)}
\end{equation}

with $precision =\frac {TP}{(TP + FP)}$ and $recall = \frac {TP}{(TP + FN)}$, where \textit{TP, FP, TN, FN} stand for True Positives, False Positives, True Negatives and False Negatives. We chose the macro-averaged F1 (i.e. employing the unweighted mean of precision and recall for both classes) since our data-set was perfectly balanced.

\subsection{Models}
As well as our novel model for joint survival time and churn probability estimation we discuss several models for disjoint estimation, learning only survival time or churn probability, in order to conduct our experiments and  compare our model with existing techniques. Furthermore, for providing a baseline comparison in our experiments we employed a mean model (MM), which generates predictions based on the average of the targets in the training set.

\subsubsection{Models for disjoint estimation}
The choice of disjoint estimation models was dictated by a series of needs: widespread usage in research and industry settings, ability to capture linear and non-linear interactions between features and most importantly capability to train on large data-sets (e.g. matrix of dimension $\approx10^6\times10^2$). Four models were employed in Experiments 1 and 2. Firstly, a variant of Regularized Regression, ElasticNet (EN) \cite{zou2005regularization}, for survival estimation and Logistic Regression (LR) for churn probability estimation. Secondly, a pair of similar Multi-Layers Perceptron Neural Networks, one tasked to perform survival time regression, MLPr, and one to perform churn classification, MLPc. We felt that given the similarities between linear models and NNs, which can be seen as a stacked version of the former with more `expressive power', the chosen algorithms constituted a natural progression in the modelling approach. For EN the best hyper-parameters were $\alpha = 0.1$ and a ratio of $0.5$ between l1 and l2 regularization. For LR an l1 regularization with $C = 0.01$. Both MLPr and MLPc employed an l2 penalty of $0.01$ and utilized a 3 layers architecture with 200, 100 and 50 hidden units. For all hidden units a $ReLU(z) = max(0, z)$ activation function was used, while an  $identity(z) = z$ and $ sigmoid(z) = \frac {1} {1 + \epsilon^{-z}}$ functions were respectively used as final activations for the MLPr and MLPc, where $z$ is a weighted sum of the hidden units of the previous layer. When training the MLP based models a small sub-set was extracted from the training set which represented 10\% of the data. This sub-set was used to evaluate convergence of the model and stop the training phase before over-fitting could occur. For both models convergence was determined if the loss did not improve for 3 epochs. The networks were trained using a batch size of 256 and optimized using the Adaptive Moment Estimation (ADAM) optimizer \cite{kingma2014adam}. Because survival time estimation is a regression task and churn prediction is classification task different loss function were used, Mean Squared Error (MSE) and Binary Cross Entropy (BCE) respectively. These are defined as:

\begin{equation}
\label{mae}
    \text{MSE}=
        \dfrac
            {1}
            {N}
            \sum\limits_{i=1}^{N}  (y_i - \hat{y_i})^2
\end{equation}
\begin{equation}
\label{bce}
    \text{BCE}=
        -\dfrac
            {1}
            {N}
        \sum\limits_{i=1}^{N}  y_i \cdot log(\hat{y_i}) + (1-y_i) \cdot log(1 - \hat{y_i})
\end{equation}

where $N$ is the size of the batch, and $\hat{y_i}$ and $y_i$ are respectively estimations provided by the model and ground truth value for the $i_{th}$ element in the batch. 

\begin{figure}[h]
\centering
\includegraphics[width=0.5\textwidth]{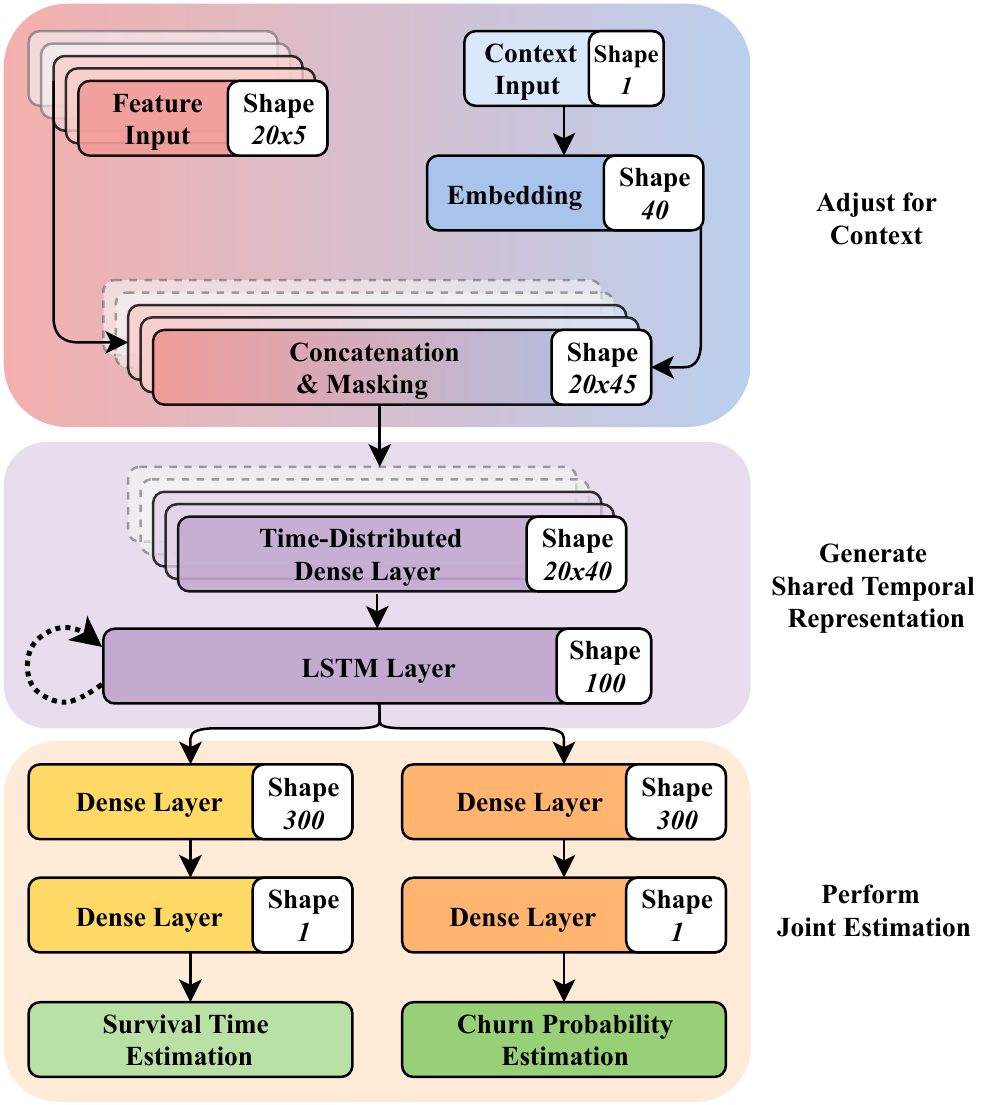}
\caption{\textbf{Bifurcating Model (BM) Architecture.} The first section learns an embedding for each game context and fuses it, via concatenation, with the feature set. The embedding allows the model to learn a rich multi-dimensional representation of the game context projecting similar games into closer points in the latent space. The second section takes these fused representation over time and models them temporally using an LSTM. The LSTM is particularly suitable because it can handle time series of different lengths and explicitly model temporal dependencies. We thought to use this part of the model for extracting a high level representation of the player state which could be used for predicting measures of future disengagement and sustained engagement. Inspired by the results of \textit{Experiment 1}, this is achieved by 'branching' two shallow NNs tasked to perform churn probability and survival time estimation.}
\label{bm_architecture}
\end{figure}

\subsubsection{Bifurcating Model for Joint Estimation}
For Experiment 3 we present a novel deep neural network architecture, loosely inspired by the winning entry in \cite{lee2018game}, for jointly estimate survival time and churn probability. This architecture, the `Bifurcating Model' (BM), is demonstrated in Fig. \ref{bm_architecture}. The model receives, as input, both a vector of unfolded features, as in Table \ref{metricsdescription}, as well as a context vector containing a numerical encoding of the game (e.g. jc3 = $[1]$, lis = $[2]$ etc.). The game context is then embedded into a vector of $l = 40$, similarly to what is done in words embedding for sentiment analysis \cite{chollet2015keras}. Differently from a one-hot encoding, this approach provides a non-sparse representation of the input while also projecting it into a multi-dimensional space where the relationships between elements become meaningful (e.g. game contexts which are similar to each other in respect to the objectives will be located closer to each other in the embedding space). Using an embedding for encoding the game contexts allows to have a representation that grows richer and richer the more categories are included into it. Obviously this would require to re-train the model whenever a new unseen context is added, practice however not just advisable but also routinely done in production. Next, the raw behavioural input and the embedded game context vector are concatenated along the temporal dimension into a single feature vector and a zero-padding re-applied where needed. At this point, a masking layer allows the model to more efficiently work with time-series of different lengths (i.e. skipping the computations for the zero-padded time-steps) and a dense layer, applied to each time step, to combine raw behavioural metrics and context in a new vector of $l = 40$. These newly obtained features are then modelled across time using a Long Short-Term Memory (LSTM) recurrent layer with $n = 100$ units. Therefore, the output of this LSTM Layer is a feature vector of $l = 100$ which is a latent representation of the input features across time and can be seen as providing a high-level representation of the behavioural state of the user during the OP. The final step of this architecture is to then take this high-level latent representation and pass it to a pair of shallow NNs, one tasked with estimating survival time and the other churn probability. These estimators are formed of a pair of densely connected layers, where the first layer has $n = 300$ units and the last has $n = 1$ units, the output of which will constitute the survival time and churn probability estimates. Like the two MLP models the BM was batch trained with a batch size of 256 until convergence using the ADAM optimizer, with learning rate adjusted through a cyclical policy \cite{smith2017cyclical, chollet2015keras}, minimizing the sum of the two losses. Similarly to the MLP models, the hidden layers used $ReLU$ as activation function whereas the two outputs units used respectively an $identity$ and $sigmoid$ functions for producing the survival time and churn probability estimates. For the survival time branch SMAPE was used as an objective function while for the churn estimation branch BCE was adopted. We applied two regularization techniques after the computations of the first layers of each shallow NN, batch normalization \cite{ioffe2015batch} and dropout \cite{srivastava2014dropout} ($rate = 0.1$). Additionally, following the intuition from \cite{gal2016dropout}, we employed dropout also at inference time for sampling from the model parameters and obtaining a distribution over the posterior so to be able to represent uncertainty in the model estimates. This was achieved by querying the model 50 times at prediction time and retaining all the produced values. When computing the performance metrics we then used the mean of the estimated values, since they roughly followed a normal distribution the mean could be seen as the value with highest probability. All the experiments were implemented in Python 3.6, with the algorithms for Experiment 1 and 2 provided by the library scikit-learn \cite{scikit-learn} and our novel BM architecture developed using Keras with Tensorflow as a back-end \cite{chollet2015keras}.

\section{Results}
We will first present results for each disjoint model as well as for a baseline model. Next we will illustrate in detail the performance of the BM model both in terms of it's raw accuracy as well as its capability to include uncertainty in it's output. Note that for all reported SMAPE results the smaller the better as it represents the error between the prediction and ground truth. Conversely, for F1 the larger the better since it measures how often the trained model made the correct classifications without false alarms. The probability threshold employed for discriminating between classes was set to 0.5.

\begin{table}[h]\centering
\caption{\textbf{Performance Baseline Mean Model}}
\label{baseperformance}
\resizebox{0.5\textwidth}{!}{
\begin{tabular}{@{}llrr@{}}
\toprule
\textbf{Game}  &\textbf{Model}                 & \textbf{SMAPE}      & \textbf{F1}       \\ \midrule
\textbf{hmg}   &\multirow{2}{*}{}               & $0.767 \pm 0.001$   & $0.500 \pm 0.003$ \\
\textbf{hms}   &                                & $0.581 \pm 0.001$   & $0.507 \pm 0.003$ \\
\textbf{jc3}   &\multirow{2}{*}{\textbf{MM}}    & $0.632 \pm 0.003$   & $0.499 \pm 0.004$ \\
\textbf{jc4}   &                                & $0.366 \pm 0.002$   & $0.499 \pm 0.001$ \\
\textbf{lis}   &\multirow{2}{*}{\textit{}}      & $0.404 \pm 0.001$   & $0.500 \pm 0.003$ \\
\textbf{lisbf} &                                & $0.244 \pm 0.002$   & $0.500 \pm 0.005$ \\ \bottomrule
\end{tabular}
}
\end{table}

\subsection{Experiment 1}
The results from the first experiment, Table \ref{collapsedperformance}, showed how all the 4 models strongly outperformed the MM baseline, Table \ref{baseperformance}, in all games, while also achieving an overall satisfying performance. Moreover we noticed how MLPr and MLPc markedly outperformed EN and LR in both churn probability and survival time estimation across all games.  

\begin{table}[h] \centering
\caption{\textbf{Performance Collapsed Format}}
\label{collapsedperformance}
\resizebox{0.5\textwidth}{!}{
\begin{tabular}{@{}llrlr@{}}
\toprule
\textbf{Game}  &\textbf{Model}                & \textbf{SMAPE}       &\textbf{Model}                & \textbf{F1}       \\ \midrule
\textbf{hmg}   &\multirow{2}{*}{}             & $0.513 \pm 0.043$    &\multirow{2}{*}{}             & $0.591 \pm 0.004$ \\
\textbf{hms}   &                              & $0.331 \pm 0.020$    &                              & $0.624 \pm 0.004$ \\
\textbf{jc3}   &\multirow{2}{*}{\textbf{EN}}  & $0.423 \pm 0.008$    &\multirow{2}{*}{\textbf{LR}}  & $0.601 \pm 0.004$ \\
\textbf{jc4}   &                              & $0.351 \pm 0.006$    &                              & $0.663 \pm 0.002$ \\
\textbf{lis}   &\multirow{2}{*}{}             & $0.287 \pm 0.004$    &\multirow{2}{*}{}             & $0.626 \pm 0.003$ \\
\textbf{lisbf} &                              & $0.239 \pm 0.003$    &                              & $0.591 \pm 0.003$ \\ \midrule

\textbf{hmg}   &\multirow{2}{*}{}             & $0.304 \pm 0.008$    &\multirow{2}{*}{}             & $0.660 \pm 0.006$ \\
\textbf{hms}   &                              & $0.241 \pm 0.007$    &                              & $0.670 \pm 0.006$ \\
\textbf{jc3}   &\multirow{2}{*}{\textbf{MLPr}}& $0.360 \pm 0.003$    &\multirow{2}{*}{\textbf{MLPc}}& $0.654 \pm 0.004$ \\
\textbf{jc4}   &                              & $0.334 \pm 0.002$    &                              & $0.678 \pm 0.004$ \\
\textbf{lis}   &\multirow{2}{*}{\textit{}}    & $0.256 \pm 0.003$    &\multirow{2}{*}{\textit{}}    & $0.664 \pm 0.003$ \\
\textbf{lisbf} &                              & $0.219 \pm 0.002$    &                              & $0.622 \pm 0.003$ \\ \bottomrule
\end{tabular}
}
\end{table}

\subsection{Experiment 2}
Following the results of Experiment 1 we tested the same modelling approaches on the unfolded version of the features, where all data points are provided rather than summary statistics. We observed a similar pattern of results, see Table \ref{unfoldedperformance}, regarding baseline and inter-models comparisons. However, it was clear that using unfolded, temporal data lead to only small improvements over the aggregated data from Experiment 1. This might be explained by the fact that the chosen modelling approaches are not explicitly designed for taking temporal structure into account, for example they have no explicit mechanics for temporal modelling such as those provided by a LSTM.

\begin{table}[h] \centering
\caption{\textbf{Performance Unfolded Format}}
\label{unfoldedperformance}
\resizebox{0.5\textwidth}{!}{
\begin{tabular}{@{}llrlr@{}}
\toprule
\textbf{Game}  &\textbf{Model}                 & \textbf{SMAPE}  &\textbf{Model}               & \textbf{F1}       \\ \midrule
\textbf{hmg}   &\multirow{2}{*}{}             & $0.545 \pm 0.024$ &\multirow{2}{*}{}             & $0.612 \pm 0.004$ \\
\textbf{hms}   &                              & $0.550 \pm 0.020$ &                              & $0.626 \pm 0.004$ \\
\textbf{jc3}   &\multirow{2}{*}{\textbf{EN}}  & $0.384 \pm 0.003$ &\multirow{2}{*}{\textbf{LR}}  & $0.607 \pm 0.003$ \\
\textbf{jc4}   &                              & $0.349 \pm 0.002$ &                              & $0.660 \pm 0.003$ \\
\textbf{lis}   &\multirow{2}{*}{}             & $0.302 \pm 0.001$ &\multirow{2}{*}{}             & $0.641 \pm 0.004$ \\
\textbf{lisbf} &                              & $0.235 \pm 0.002$ &                              & $0.578 \pm 0.003$ \\ \midrule
\textbf{hmg}   &\multirow{2}{*}{}             & $0.293 \pm 0.004$ &\multirow{2}{*}{}             & $0.683 \pm 0.005$ \\
\textbf{hms}   &                              & $0.226 \pm 0.004$ &                              & $0.682 \pm 0.004$ \\
\textbf{jc3}   &\multirow{2}{*}{\textbf{MLPr}}& $0.360 \pm 0.003$ &\multirow{2}{*}{\textbf{MLPc}}& $0.643 \pm 0.004$ \\
\textbf{jc4}   &                              & $0.331 \pm 0.002$ &                              & $0.681 \pm 0.003$ \\
\textbf{lis}   &\multirow{2}{*}{\textit{}}    & $0.256 \pm 0.002$ &\multirow{2}{*}{\textit{}}    & $0.673 \pm 0.005$ \\
\textbf{lisbf} &                              & $0.218 \pm 0.001$ &                              & $0.627 \pm 0.003$ \\ \bottomrule
\end{tabular}
}
\end{table}

\subsection{Experiment 3}
\begin{figure}[h]

  \centering
  \includegraphics[width=8.5cm]{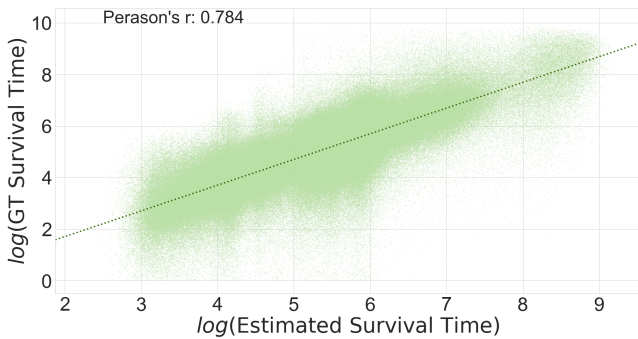}
  \caption{\textbf{Performance of the BM on survival task}. The scatter plot shows the relationship between the survival estimates provided by the BM and the ground truth values. Since the relationship is evaluated on the $log$ of both variables, due to the presence extreme outliers in the ground truth, this acts as mostly as a qualitative complement to the more reliable SMAPE measure.}
  \label{perfsurv}
  \end{figure}

\begin{figure}[h]
  \centering
  \subfloat[Survival Estimations]{\includegraphics[ width=8.5cm]{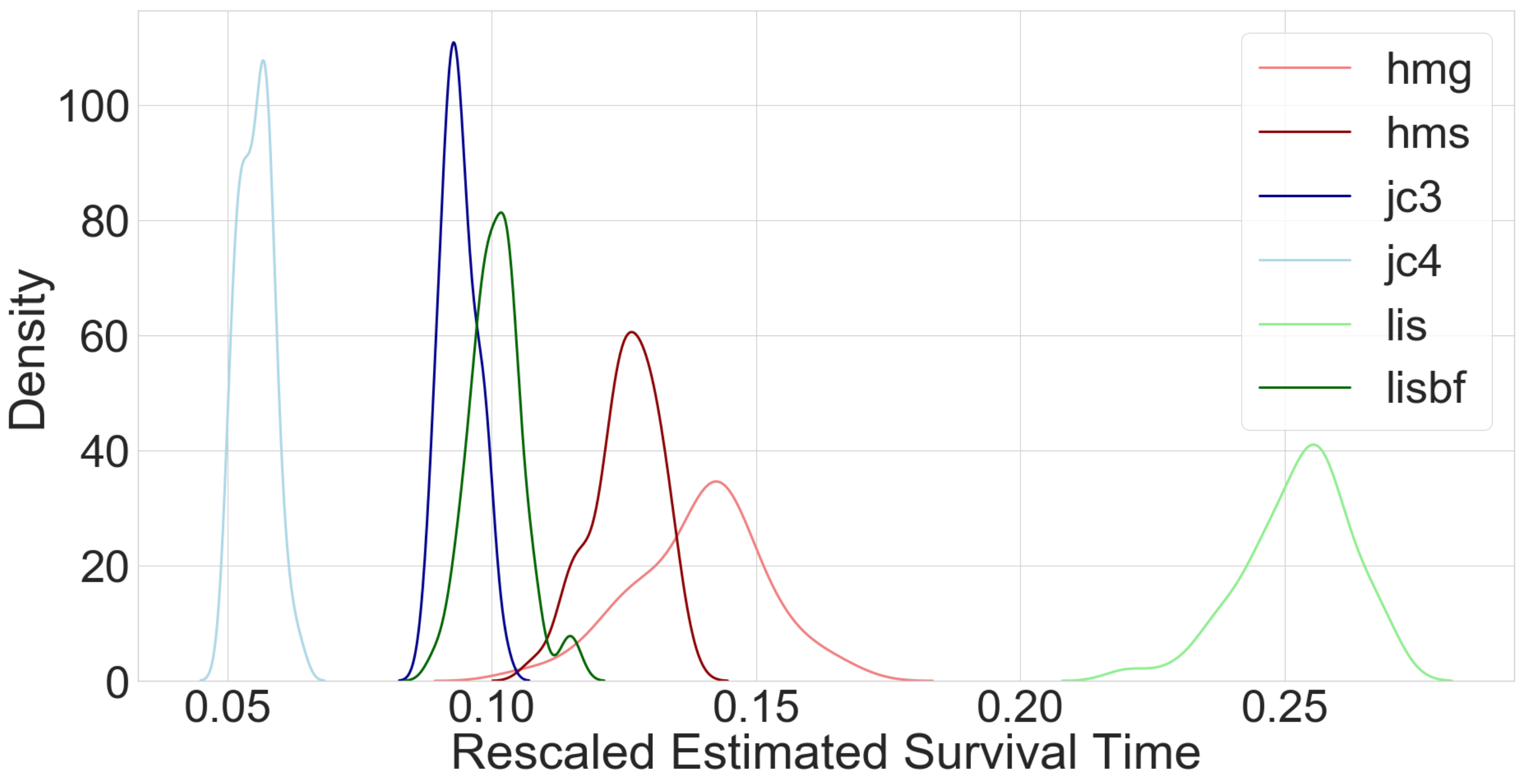}\label{fig:densurv}}
  \hfill
  \subfloat[Churn Probability Estimation]{\includegraphics[ width=8.5cm]{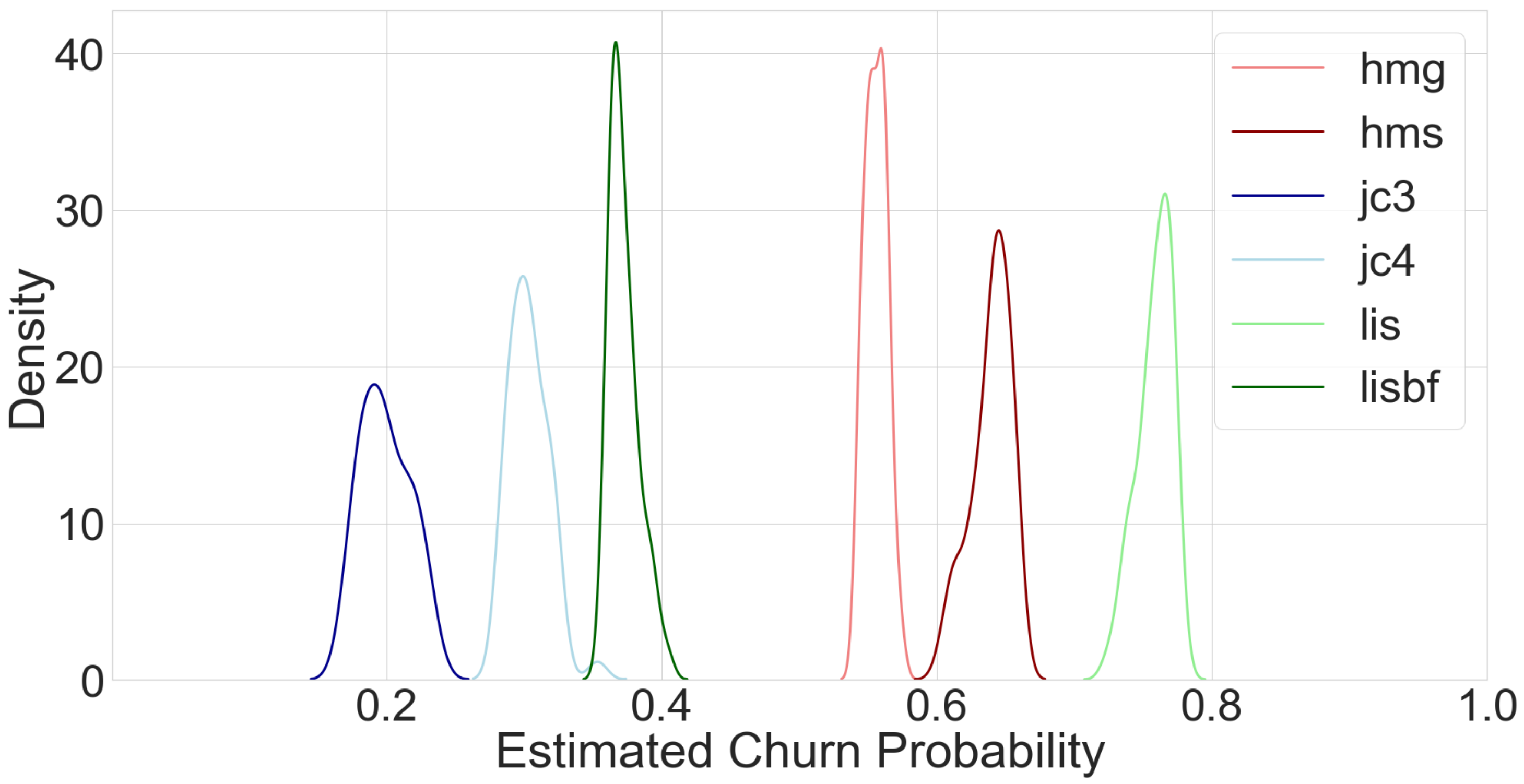}\label{fig:denchurn}}
  \caption{\textbf{Distribution of the BM estimates for six random users, one for each game}. For better comparison the survival estimates are re-scaled game-wise in the range 0 to 1. The highest density point in the distribution represents the most probable estimated value (i.e. the actual prediction),  while the area under the curve instead can be seen as measure of uncertainty (i.e. how confident is the model in its prediction).}
  \label{distestimations}
\end{figure}

Informed by the results of Experiment 1 and 2, we proceeded in evaluating the performance of our BM, Table \ref{bifurcatingperformance}, on the unfolded data. We observed how our model achieved a modest but consistent improvements in both churn probability and survival time estimation in all game contexts compared to the previous best model (MLPr and MLPc). From a visual inspection of Figure \ref{perfsurv} we can see the presence of a positive linear relationship between estimated and ground truth survival time (indicative of accordance between the two), with a roughly even distribution of error along the entire range of values. In Table \ref{confusionmatrix} we can observe how the model performance is evenly split across the two classes highlighting similar levels of precision and recall. Finally, observing the density plots in Figure \ref{fig:densurv} and \ref{fig:denchurn} we can see how the model was able to encode different levels of uncertainty through the distribution's variance of estimated values.

\begin{table}[h] \centering
\caption{\textbf{Performance Bifurcating Model}}
\label{bifurcatingperformance}
\resizebox{0.5\textwidth}{!}{
\begin{tabular}{@{}llrr@{}}
\toprule
\textbf{Game}  &\textbf{Models}               & \textbf{SMAPE}      & \textbf{F1}       \\ \midrule
\textbf{hmg}   &\multirow{2}{*}{}             & $0.275 \pm 0.001$   & $0.693 \pm 0.002$ \\
\textbf{hms}   &                              & $0.200 \pm 0.001$   & $0.701 \pm 0.003$ \\
\textbf{jc3}   &\multirow{2}{*}{\textbf{BM}}  & $0.344 \pm 0.003$   & $0.671 \pm 0.005$ \\
\textbf{jc4}   &                              & $0.325 \pm 0.002$   & $0.685 \pm 0.002$ \\
\textbf{lis}   &\multirow{2}{*}{}             & $0.246 \pm 0.002$   & $0.688 \pm 0.003$ \\
\textbf{lisbf} &                              & $0.208 \pm 0.001$   & $0.645 \pm 0.003$ \\ \bottomrule
\end{tabular}
}
\end{table}

\begin{table}[h] \centering
\caption{\textbf{Performance of the BM on churn task.} Here the diagonal shows the \% of correctly predicted users for each label across all games.}
\label{confusionmatrix}
\begin{tabular}{llcc}
\toprule
 & & \multicolumn{2}{c}{\textbf{Estimation}} \\ \cmidrule(lr){3-4}
 & & \parbox[c]{1.5cm}{Churner} & \parbox[c]{1.5cm}{Non-Churner} \\ \midrule
\multirow{2}{*}{\rotatebox{90}{\parbox[c]{1cm}{\textbf{Ground  Truth}}}} 
&\multirow{2}{*}{Churner}  & \cellcolor{DarkOliveGreen3!69}   &  \cellcolor{DarkOliveGreen3!31}   \\ 
&&\multirow{-2}{*}{\cellcolor{DarkOliveGreen3!69}0.69}& \multirow{-2}{*}{\cellcolor{DarkOliveGreen3!31}0.31}\\
&\multirow{2}{*}{Non-Churner}  &  \cellcolor{DarkOliveGreen3!33}   &  \cellcolor{DarkOliveGreen3!66}   \\ 
&  &  \multirow{-2}{*}{\cellcolor{DarkOliveGreen3!33} 0.33} & \multirow{-2}{*}{\cellcolor{DarkOliveGreen3!66} 0.66}\\
\bottomrule
\end{tabular}
\end{table}

%%%%%%%%%%%%%%%%%%%%%%%%%%%%%%%%%%%%%%%%%%%%%%%%%%%%%%%%%%%%%%%%%%%%%%%%%%%%%%%%%%%%%%%%%%%%%%

\section{Discussion, limitations and future work}
The results of our experiments highlight how employing metrics indicative of behavioural activity in early user-game interactions allowed our model to estimate proxy measures of future disengagement and sustained engagement. This suggests that the early user-game interactions might be relevant for characterizing long-term engagement as well as that measures of behavioural activity could be a useful index for its inference \cite{milovsevic2017early, mirza2013does}. We also found how the use of non-parametric models, able to capture non-linear interactions between features provided substantial improvements in estimating proxy measures of engagement when compared to simpler, although computationally cheaper, parametric ones. We also show that including temporal structure explicitly provides a slight edge over metrics representations which are collapsed over time, moreover we noticed that this improvement is more pronounced and consistent when employing approaches that explicitly model temporality, i.e. the BM. This is in accordance with the aforementioned theoretical formalization of engagement as a dynamic process rather than a static construct \cite{o2008user}. Finally the visual representation of the performance of the BM highlighted how the proposed methodology generalizes well when trying to predict survival time and churn probability as well as successfully incorporating measures of uncertainty in its estimations. 

While the work presented here crosses various game genres, it does not include all the major ones (e.g. multi-player titles). Moreover, despite acknowledging the complexity of the chosen estimation task, better model performance would have been desirable. Finally, the heavy dependence on a supervised approach for learning the context embedding and the inability to fully exploit the LSTM potential (i.e. our time series were at maximum 20 steps long) limited the potential of our approach. Future work will try to improve on these drawbacks considering more game genres, integrating approaches for learning context in an unsupervised way and taking into consideration longer streams of sessions. We will also try to explicitly model the contribution of elements external to the game environment for taking into account the impact of real-world factors (e.g. day of the week or time of the day).

\section{Acknowledgments}
We would like to thank Ivan Bravi, Simon Demediuk and Peter York for their invaluable insights provided during the realization of this work and the \textit{Square Enix Limited} Analytics and Insight Team for the continuous support provided.

\bibliographystyle{IEEEtran}
\bibliography{main}
\end{document}